\documentclass[10pt,twocolumn,letterpaper]{article}

\usepackage{cvpr}              

\usepackage[toc,page]{appendix}
\usepackage[accsupp]{axessibility}

\usepackage{graphicx}

\usepackage{array}
\usepackage{tabularx}

\usepackage{enumitem}

\usepackage[utf8]{inputenc} 
\usepackage[T1]{fontenc}    
\usepackage{url}            
\usepackage{booktabs}       
\usepackage{amsfonts}       
\usepackage{nicefrac}       
\usepackage{microtype}      
\usepackage{xcolor}         

\usepackage{algorithm}
\usepackage{algpseudocode}
\usepackage{amsmath,amssymb,mathtools,amsthm,bm,etoolbox}
\usepackage{bbm, dsfont}
\usepackage{cancel, ulem}

\newtheorem{theorem}{Theorem}

\newtheorem{corollary}{Corollary}
\newtheorem{prop}{Proposition}
\newtheorem{assumption}{Assumption}

\newcommand{\sgn}{\text{sgn}}
\newcommand{\grad}{\nabla}

\newcommand{\EE}{\mathbb{E}}

\newcommand{\gradd}{\grad_{\delta}}

\DeclareMathOperator*{\argmax}{argmax}
\DeclarePairedDelimiter\floor{\lfloor}{\rfloor}
\newcommand{\nn}{\|}

\newcommand{\loss}{\mathcal{L}}

\newcommand{\eps}{\epsilon}
\newcommand{\diam}{\text{diam}}

\newcommand{\fwadapt}{\textsc{FW-AT-Adapt }}
\newcommand{\fwk}[1]{\textsc{FW(#1)}}
\newcommand{\pgdk}[1]{\textsc{PGD(#1)}}
\newcommand{\pgdat}{\textsc{PGD-AT}}
\newcommand{\pgdatk}[1]{\textsc{PGD(#1)-AT}}

\makeatletter
\newcommand{\printfnsymbol}[1]{%
  \textsuperscript{\@fnsymbol{#1}}%
}
\makeatother

\usepackage[pagebackref,breaklinks,colorlinks]{hyperref}

\usepackage[capitalize]{cleveref}
\crefname{section}{Sec.}{Secs.}
\Crefname{section}{Section}{Sections}
\Crefname{table}{Table}{Tables}
\crefname{table}{Tab.}{Tabs.}

\def\cvprPaperID{9075} 
\def\confName{CVPR}
\def\confYear{2022}

\begin{document}

\title{Understanding and Increasing Efficiency of Frank-Wolfe Adversarial Training}

\author{Theodoros Tsiligkaridis\thanks{Equal contributions.},\quad Jay Roberts\printfnsymbol{1}\\
Massachusetts Institute of Technology Lincoln Laboratory\\
Lexington, MA 02421\\
{\tt\small ttsili@mit.edu},\quad {\tt\small jay.roberts@ll.mit.edu}
}


\maketitle

\begin{abstract}
Deep neural networks are easily fooled by small perturbations known as adversarial attacks. Adversarial Training (\textsc{AT}) is a technique that approximately solves a robust optimization problem to minimize the worst-case loss and is widely regarded as the most  effective defense against such attacks. Due to the high computation time for generating strong adversarial examples in the \textsc{AT} process, single-step approaches have been proposed to reduce training time. However, these methods suffer from catastrophic overfitting where adversarial accuracy drops during training, and although improvements have been proposed, they increase training time and robustness is far from that of multi-step \textsc{AT}. We develop a theoretical framework for adversarial training with \textsc{FW} optimization (\textsc{FW-AT}) that reveals a geometric connection between the loss landscape and the distortion of $\ell_\infty$ \textsc{FW} attacks (the attack's $\ell_2$ norm). Specifically, we analytically show that high distortion of \textsc{FW} attacks is equivalent to small gradient variation along the attack path. It is then experimentally demonstrated on various deep neural network architectures that $\ell_\infty$ attacks against robust models achieve near maximal $\ell_2$ distortion, while standard networks have lower distortion. Furthermore, it is experimentally shown that catastrophic overfitting is strongly correlated with low distortion of \textsc{FW} attacks. This mathematical transparency differentiates \textsc{FW} from the more popular Projected Gradient Descent (\textsc{PGD}) optimization. To demonstrate the utility of our theoretical framework we develop \fwadapt, a novel adversarial training algorithm which uses a simple distortion measure to adapt the number of attack steps during training to increase efficiency without compromising robustness. \fwadapt provides training time on par with single-step fast AT methods and improves closing the gap between fast AT methods and multi-step \pgdat with minimal loss in  adversarial accuracy in white-box and black-box settings.
\end{abstract}

\section{Introduction}

Deep neural networks (DNN) achieve excellent performance across various domains \cite{LeCun:2015}. 
As these models are deployed across industries (e.g., healthcare or autonomous driving), concerns of robustness and reliability become increasingly important. Several organizations have identified important principles of
artificial intelligence (AI) that include the notions of reliability and transparency \cite{google:blog, microsoft:blog, dod:blog}.

One issue of large capacity models such as DNNs is that small, carefully chosen input perturbations, known as adversarial perturbations, can lead to incorrect predictions~\cite{goodfellow:2015}. Various enhancement methods have been proposed to defend against adversarial perturbations \cite{kurakin:2017, ros2018:gradreg, madry2018:at, md2019:cure}. One of the best performing algorithms is adversarial training (\textsc{AT}) \cite{madry2018:at}, which is formulated as a robust optimization problem \cite{Shaham:2018}. Computation of optimal adversarial perturbations is NP-hard \cite{Weng:2018} and approximate methods are used to solve the inner maximization. The most popular approximate method that has been proven to be successful is projected gradient descent (\textsc{PGD}) \cite{Croce:2020}.  
Frank-Wolfe (\textsc{FW}) optimization has been recently proposed in \cite{Chen:2020} and was shown to effectively fool standard networks with less distortion, and can be efficiently used to generate sparse counterfactual perturbations to explain model predictions and visualize principal class features \cite{Roberts:MLVis:2021}.

Since \textsc{PGD} has proven to be the main algorithm for adversarially robust deep learning, reducing its high computational cost without sacrificing performance, i.e. fast adversarial training, is a primary issue. Various methods have been proposed based on using a single \textsc{PGD} step, known as Fast Gradient Sign Method (FGSM) \cite{Wong:2020} but fail for large perturbations. \cite{Wong:2020} identified that FGSM-based training achieves some robustness initially during training but robustness drastically drops within an epoch, a phenomenon known as \textit{catastrophic overfitting} (CO). While some methods have been proposed to ameliorate this problem \cite{Andriushchenko:2020, KimLeeLee:2021, Shafahi:2019}, the training time suffers as a result and/or robustness is not on par with multi-step PGD-AT.

\begin{figure}
    \centering
    \includegraphics[width=0.4 \textwidth]{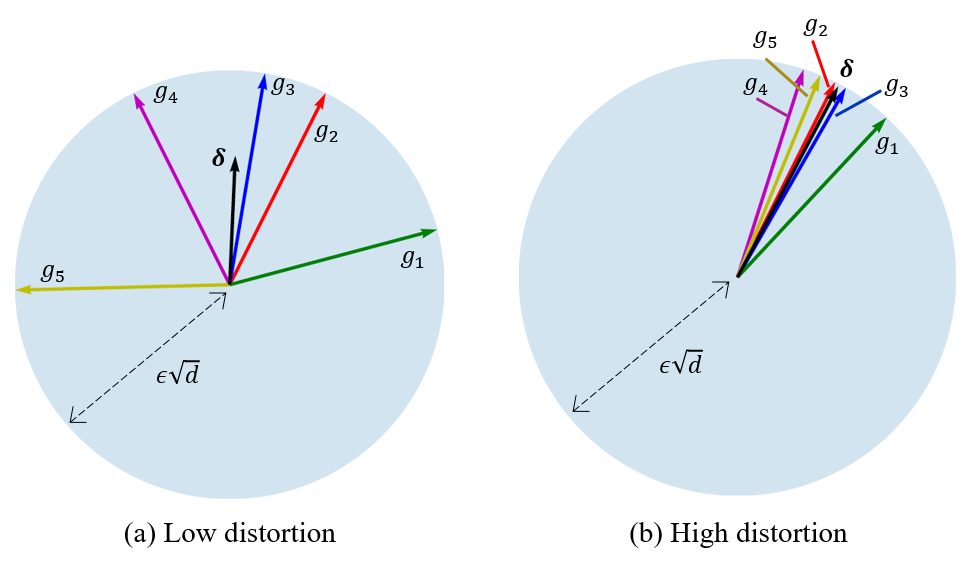}
    \caption{ Low (high) distortion of attacks $\delta$ (black) is equivalent to high (low) angular spread of signed gradients $g_l = \epsilon \cdot \sgn(\grad_\delta \loss(x+\delta_l,y))$ (all with $\nn g_l \nn_2=\epsilon \sqrt{d}$) computed over $K$ steps along attack path ($K=5$ here). Proposition \ref{prop:FW_adv_pert} expresses FW adversarial perturbations $\delta$ as convex combinations of signed gradients along the attack path. This core concept is quantified in Theorem \ref{thm:FW_distortion}, and forms the basis for the development of adaptive adversarial training algorithm presented in Section \ref{sec:FW_AT_Adapt}. }
    \label{fig:distortion_concept}
\end{figure}

\begin{figure}
    \centering
    \includegraphics[width=0.45 \textwidth]{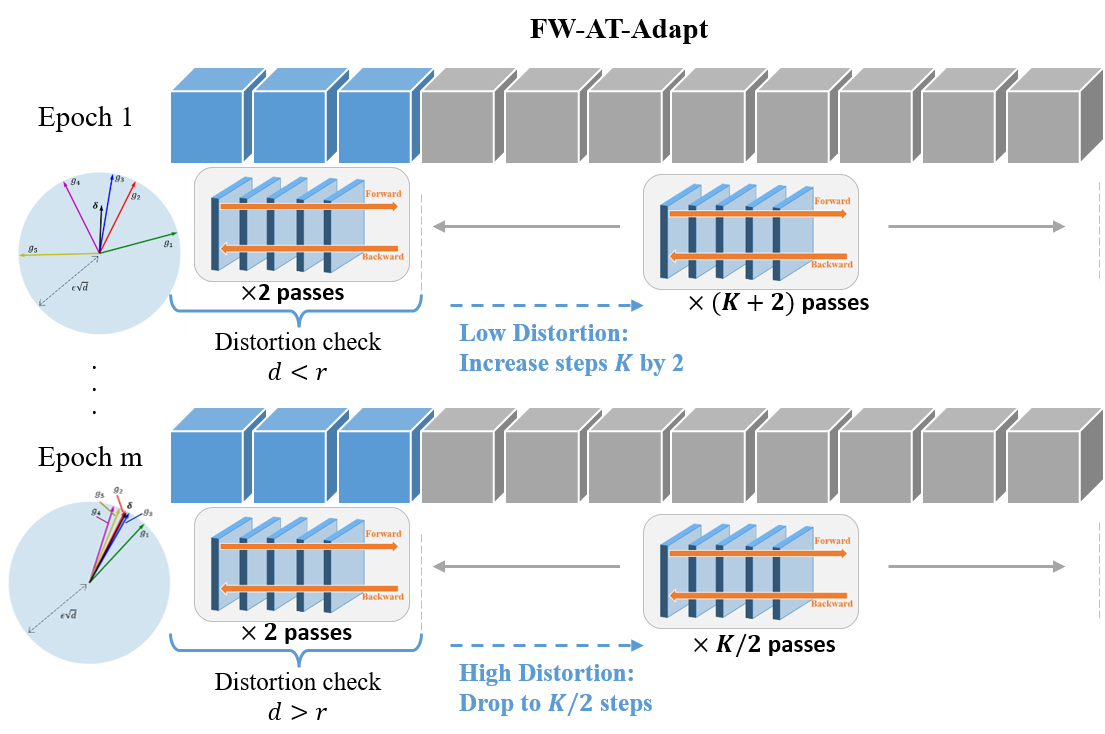}
    \caption{ Illustration of concept behind \textsc{FW-AT-Adapt} training algorithm. At each epoch, distortion $d$ is monitored across the first $B_m$ batches (here $B_m=3$). If the average distortion is less than a threshold $r$, the current number of steps, $K$, is increased by $2$, to $K+2$, (and if it is higher than a threshold $r$, the current number of steps, $K$, is dropped by a factor of $2$, to $K/2$) for the remaining batches in the epoch. This process is repeated until convergence and reduces the training time of adversarial training without sacrificing robustness. Theorems \ref{thm:FW_distortion_grad} and \ref{thm:weight_update_bound} provide stability guarantees for robust model weight updates in the high-distortion regime. }
    \label{fig:FW_AT_Adapt_concept}
\end{figure}

In this paper, we use the Frank-Wolfe optimization to derive a relationship between the $\ell_2$ norm of $\ell_{\infty}$ adversarial perturbations (distortion) and the geometry of the loss landscape (see Fig. \ref{fig:distortion_concept}). Using this theory and empirical studies we show that this distortion can be used as a signal for CO and propose a fast adversarial training algorithm based on an adaptive Frank-Wolfe adversarial training (\fwadapt) method (see Fig. \ref{fig:FW_AT_Adapt_concept}). This method yields training times on par with single step methods without suffering from CO, outperforms numerous single step methods, and begins to close the gap between fast adversarial training methods and multi-step \textsc{PGD} adversarial training.

Our main contributions are summarized below:

\begin{itemize}[noitemsep,topsep=0pt]

\item We demonstrate empirically that \textsc{FW} attacks against robust models achieve near-maximal distortion across a variety of network architectures.

\item We empirically show that distortion of \textsc{FW} attacks, even with only 2 steps, are strongly correlated with catastrophic overfitting.

\item Theoretical guarantees are derived that relate distortion of \textsc{FW} attacks to the gradient variation along the attack path and which imply that high distortion attacks computed with several steps result in diminishing increases to the loss.



\item Inspired by the connection between distortion and attack path gradient variation, we propose an adaptive step Frank-Wolfe adversarial training algorithm, \fwadapt, which achieves superior robustness/training-time tradeoffs compared to single-step AT and closes the gap between such methods and multi-step AT variants when evaluated against strong white- and black-box attacks.

\end{itemize}


\section{Background and Previous Work}
Consider $(x_i,y_i)\sim \mathcal{D}$ pairs of data examples drawn from distribution $\mathcal{D}$. The labels  span $C$ classes. 
The neural network function $f_\theta(\cdot)$ maps input features into logits, where $\theta$ are the  model parameters. 
The predicted class label is given by $\hat{y}(x)=\arg \max_c f_{\theta,c}(x)$.

\noindent \textbf{Adversarial Training.} The prevalent way of training classifiers is through empirical risk minimization (ERM):

\begin{equation} \label{eq:ERM}
    \min_\theta \EE_{(x,y)\sim \mathcal{D}} [\loss(x,y;\theta)]
\end{equation}

where $\loss$ is the usual cross-entropy loss.
Adversarial robustness for a classifier $f_\theta$ is defined with respect to a metric, here chosen as the $\ell_p$ metric associated with the ball $B_p(\epsilon)=\{\delta:\nn \delta \nn_p\leq \epsilon\}$, as follows. A network is said to be robust to adversarial perturbations of size (or strength) $\epsilon$ at a given input example $x$ iff $\hat{y}(x)=\hat{y}(x+\delta)$ for all $\delta \in B_p(\epsilon)$, i.e., if the predicted label does not change for all perturbations of size up to $\epsilon$. 
Training neural networks using the ERM principle (\ref{eq:ERM}) gives high accuracy on test sets but leaves the network vulnerable to adversarial attacks. 

One of the most popular and effective defenses is adversarial training (AT) \cite{madry2018:at} which, rather than using the ERM principle, minimizes the adversarial risk
\begin{equation} \label{eq:adv_risk}
    \min_\theta \EE_{(x,y)\sim \mathcal{D}} 
        \left[ 
            \max_{\delta \in B_p(\epsilon)}  \loss(x+\delta,y;\theta) \right].
\end{equation}

This framework was extended in the TRADES algorithm \cite{Zhang:2019} which proposes a modified loss function that captures the clean and adversarial accuracy tradeoff. Local Linearity Regularization (LLR) \cite{qin:2019} uses an analogous approach where the adversary is chosen to maximally violate the local linearity based on a first order approximation to \eqref{eq:adv_risk}.

To construct their adversarial attacks at a given input $x$ these defenses  use Projected Gradient Descent (\textsc{PGD}) to approximate the constrained inner maximization using a fixed number of steps. \textsc{PGD} computes adversarial perturbations using the iterative updates:
\begin{equation} \label{eq:pgd}
    \delta_{k+1} = P_{B_p(\epsilon)}\left( \delta_{k} + \alpha \grad_\delta \loss(x+\delta_{k},y;\theta) \right)
\end{equation}
where $P_{B_p(\epsilon)}(z)=\arg \min_{u\in B_p(\epsilon)} \nn z - u \nn_2^2$ is the orthogonal projection onto the constraint set. We refer to adversarial training using $K$ step \textsc{PGD} as \pgdatk{K}.

The computational cost of this method is dominated by the number of steps used to approximate the inner maximization, since a $K$ step \textsc{PGD} approximation to the maximization involves $K$ forward-backward propagations through the network. While using fewer \textsc{PGD} steps can lower this cost, these amount to weaker attacks which can lead to gradient obfuscation \cite{Papernot:2017, Uesato:2018}, a phenomenon where networks learn to defend against gradient-based attacks by making the loss landscape highly non-linear, and less robust models. Many defenses have been shown to be evaded by newer attacks, while adversarial training has been demonstrated to maintain state-of-the-art robustness \cite{Athalye:2018, Croce:2020}. This performance has only been improved upon via semi-supervised methods \cite{Carmon:2019, Uesato:2019}.


\noindent \textbf{Fast Adversarial Training.} Various fast adversarial training methods have been proposed that use fewer \textsc{PGD} steps. In \cite{Wong:2020} a single step of \textsc{PGD} is used, known as Fast Gradient Sign Method (\textsc{FGSM}), together with random initializations within the constraint ball, called \textsc{FGSM-Rand}, and achieves a good level of robustness at a lower computational cost. In \cite{Andriushchenko:2020} it was shown that the random initialization of \textsc{FGSM-Rand} can improve the linear approximation quality of the inner maximization, but still suffers from catastrophic overfitting (CO), a phenomenon whereby the model achieves strong robustness to the weaker training attack but is completely fooled by stronger multi-step attacks, which the authors overcome via a regularizer which penalizes gradient misalignment and we refer to as \textsc{FGSM-GA}. However, the double backpropagation needed for this method resulted in significantly higher training times than \textsc{FGSM-Rand}. 
In \cite{KimLeeLee:2021} the authors demonstrate that CO was the result of nonlinearities in the loss which resulted in higher losses on the interior of the ray connecting $x$ and $x+\delta$, thereby making them more susceptible to multi-step attacks. To combat this, the authors adapted the size of the FGSM step by sampling along this ray which we refer to as \textsc{FGSM-Adapt}.
The free adversarial training method of \cite{Shafahi:2019}, \textsc{Free-AT}, recycles the gradient information computed when updating model parameters through minibatch replay. The robustness performance of all these single-step AT variants lag far behind that of the multi-step \pgdat.


Additional methods adapt the number of steps used to approximate adversarial attacks. Curriculum learning \cite{CaiCat18} monitors adversarial performance during training and increases the number of attack steps as performance improves. Improving on this work the authors in \cite{WangFOSC:2019} use a Frank-Wolfe convergence criterion 
to adapt the number of attack steps at a given input. Both of these methods use \textsc{PGD} to generate adversarial examples and do not report improved training times.

\noindent \textbf{Frank-Wolfe Adversarial Attack.} The Frank-Wolfe (\textsc{FW}) optimization algorithm has its origins in convex optimization though recently has been shown to perform well in more general settings \cite{FW:1956, jaggi13}. The method first optimizes a linear approximation to the original problem, called a Linear Maximization Oracle (\textsc{LMO})
\begin{equation*} 
    \textsc{LMO} = \bar{\delta}_k 
        = 
            \argmax_{\delta\in B_p(\epsilon)} 
                \left<\delta, \grad_\delta \loss(x+\delta_k,y) \right>
                .
\end{equation*}
%
%
%
After calling the LMO, \textsc{FW} 
takes a step using a convex combination with the current iterate, 
%
$\delta_{k+1} = \delta_k + \gamma_k (\bar{\delta}_{k}-\delta_k)$
where $\gamma_k \in [0,1]$ is the step size. Optimizing step sizes can be found at additional computational cost;
however, in practice an effective choice is is $\gamma_k = c/(c+k)$ for some $c\geq 1$. 

The \textsc{FW} sub-problem can be solved exactly for any $\ell_p$ and the optimal $\bar{\delta}_k$ is given component-wise by $ \bar{\delta}_{k,i} = \epsilon \ \phi_p(\grad \loss_{k,i})$,
where
\begin{equation} 
\label{eq:lp opt s}
    \phi_p(\grad \loss_{k,i}) 
        =  \ \sgn(\grad \loss_{k,i}) \ 
            \begin{cases}
                e_i^{i_k^*}, & p=1 \\
                \frac{|\grad \loss_{k,i}|^{q/p}}{\nn \grad \loss^k \nn_q^{q/p}}, & 1<p<\infty \\
                1, & p=\infty \\
\end{cases},
\end{equation}
$\grad \loss = \gradd \loss(x+\delta_k,y)$, and $1/p+1/q=1$. For $p=1$, $i_k^*=\argmax_{i}|\grad \loss_{k,i}|$ and $e^{i_k^*}$ is equal to $1$ for the $i_k^*$-th component and zero otherwise. \textsc{FW} does not require a projection onto the $\ell_p$ ball which is non-trivial for $p$ not in $\{2,\ \infty \}$. For the special case of $\ell_\infty$ attacks, the optimal solution becomes the the Fast Gradient Sign Method (FGSM) ~\cite{goodfellow:2015}.

\begin{algorithm}[H]
    \caption{FW-Attack$(x,y; K, \gamma_k,  p)$
    }
    \label{alg:FW-AA}
    \begin{algorithmic}
        \State{\bfseries Input: }{%
        Model $f_\theta$, input batch $(x,y)$, max perturbation $\epsilon$, step schedule $\gamma_k$, steps $K$.
        }\\
        $\delta = 0$ 
        \For{ $0 \leq k <K$ }
            
            
            \State{}{$\delta = (1-\gamma_k)\delta + \gamma_k \ \epsilon \ \phi_p(\gradd \loss(f_\theta(x + \delta)$))}
        
        \EndFor
    \State{\bfseries Return: $\delta$}

    \end{algorithmic}
\end{algorithm}

\noindent \textbf{Our contributions.} We present Frank Wolfe Adversarial Training (\textsc{FW-AT}) which replaces the \textsc{PGD} inner optimization with a Frank-Wolfe optimizer. \textsc{FW-AT} achieves similar robustness as its \textsc{PGD} counterpart. Using a closed form expression for the \textsc{FW} attack path, we derive a geometric relationship between distortion of the attack and loss gradient variation along the attack path. This key insight leads to a simple modification of \textsc{FW-AT} where the step size at each epoch is adapted based on the $\ell_2$ distortion of the attacks and is shown to reduce training time while providing strong robustness without suffering from catastrophic overfitting.

Although our work shares some aspects with \textsc{FGSM-GA} and \textsc{FGSM-Adapt} it has several distinguishing features. Firstly, both methods are variants on FGSM which attempt to fix CO. The former by penalizing gradient misalignment and the latter by sampling steps along the FGSM direction. Our method takes multiple steps which allows it to reach points near the original FGSM direction and so avoids CO. Moreover, via our distortion analysis we show that our multi-step method can both monitor and regularize gradient variation {\bf all while simultaneously using these attacks for adversarial training}. This efficient use of multi-step attacks allows us to obtain superior robustness training time tradeoffs than either of these prior methods.

\section{Distortion of Frank-Wolfe Attacks}

Though all $\ell_p$ attacks must remain in $B_p(\epsilon)$ their $\ell_q$ norms, for $q\neq p$, can be quite different. 
This is referred to as distortion and in particular for $\ell_\infty$ attacks we are interested in insights $\ell_2$ distortion can give us into the behavior of the attack and \textsc{FW-AT}. 
In this setting, the maximal distortion possible of a $d$ dimensional input is $\eps \sqrt{d}$, and we refer to $\|\delta\|_2 / (\eps \sqrt{d})$ as the distortion ratio (or simply distortion) of the attack $\delta$. In this section, we demonstrate empirically the connection between distortion and robustness and then derive theoretical guarantees on loss gradient variation based on distortion bounds.


\begin{figure}[!tbp]
  \centering
  \includegraphics[width=0.45 \textwidth]{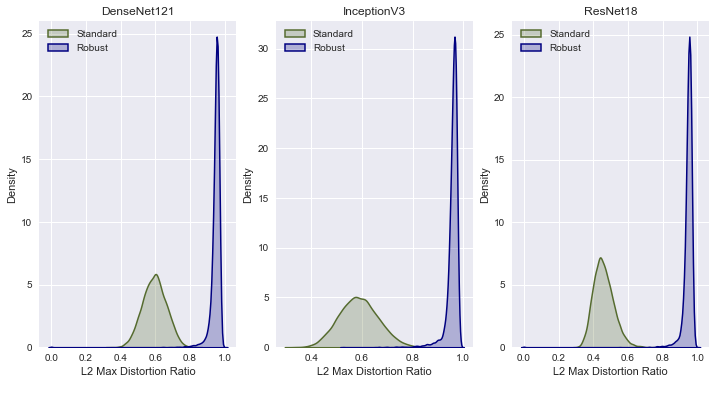}
  \caption{ Kernel density estimates of the distribution of $\nn\delta\nn_2 / (\eps \sqrt{d})$ 
  attacks computed using \fwk{20} with $\epsilon=8/255$ against standard and robust models across three architectures.  }
  \label{fig:distortion}
\end{figure}


\subsection{FW Attacks Against Robust Models are Highly Distorted}
Due to its exploitation of constraint convexity one may expect $\ell_\infty$ \textsc{FW}-attacks to remain near the interior and thus have low distortion. This was observed for standard models in \cite{Chen:2020} but robust models were not considered. Here we analyze the distortion ratio of \fwk{20} for $\ell_\infty$ constrained attacks with radius $\epsilon = 8/255$ on three architectures trained with ERM (Eq. \ref{eq:ERM}) and\pgdatk{10} (Eq. \ref{eq:adv_risk}) on CIFAR-10. 

Figure \ref{fig:distortion} shows that, while the adversarial perturbations of standard models have small distortion, robust models produce attacks that are nearly maximally distorted. In both cases attacks are near maximal in $\ell_\infty$ norm. 
This phenomenon occurs across three different architectures and is further supported by our theory below. We note that for \textsc{PGD} attacks, the distortion ratio can be trivially maximized for large step size $\alpha$, and thus this connection between distortion and robustness does not exist for \textsc{PGD} optimization. 

\subsection{Catastrophic Overfitting is Signaled by Distortion Drops} \label{sec:CO}

\begin{figure}
    \centering
    \begin{subfigure}[b]{0.43\textwidth}
    
        \centering
    \includegraphics[width=0.95\textwidth]{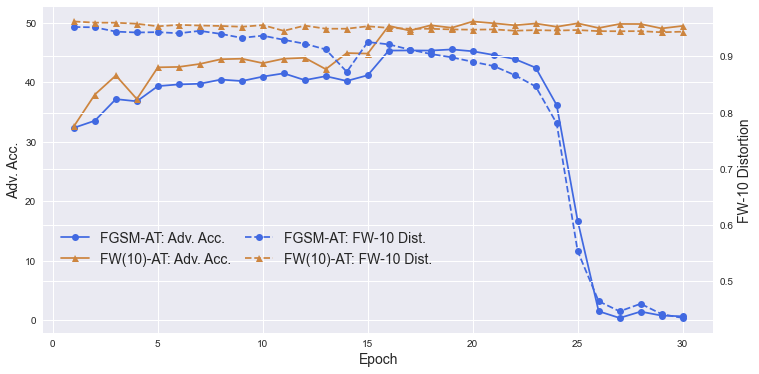}
    \caption{Adversarial accuracy and Distortion during training for \textsc{FGSM} and \textsc{\fwk{10}} adversarially trained models}
    \label{fig: dist over time}
    \end{subfigure}
    
    \begin{subfigure}[b]{0.3\textwidth}
        \centering
        \includegraphics[width=0.95\textwidth]{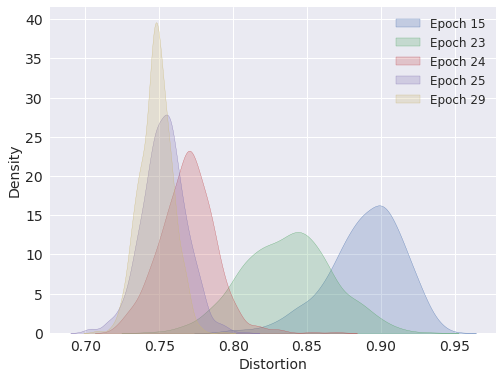}
        \caption{ Kernel density estimate of distortion of \fwk{2} attacks for \textsc{FGSM}-trained model}
        \label{fig: FW-2 Dist}
    \end{subfigure}
    
    \caption{(a) Adversarial accuracy against \pgdk{10} attacks (Blue) and average distortion (Tan) of \fwk{10} attacks on CIFAR-10 validation set for \textsc{FGSM} and \fwk{10} adversarial training at $\epsilon=8/255$. Steep drops in adversarial accuracy signaling catastrophic overfitting (CO) are accompanied by drops in distortion. (b) Kernel density estimates of distortion of a \fwk{2} attack. As CO occurs the distribution of distortion shifts and tightens on low values.}
    \label{fig:dist training}
\end{figure}

Many fast AT methods rely on a single gradient step which can lead to catastrophic overfitting (CO), a phenomenon where the model's performance against multi-step attacks converges to a high value but then suddenly plummets. This indicates that the model has overfit its weights to the single step training attack.
We demonstrate this by training with FGSM at strength $\epsilon=8/255$ for 30 epochs and plotting its validation accuracy against \pgdk{10} attacks and average distortion computed with \fwk{10} attacks. Figure \ref{fig: dist over time} demonstrates the FGSM's descent into CO, where we observe that the drop in adversarial accuracy is mirrored by a drop in the distortion of the multi-step attack. In the case of FW(10)-AT the distortion remains high throughout training.

In Figure \ref{fig: FW-2 Dist} we show this behavior is present even when evaluated against the weaker FW(2) attack. Here we plot the kernel density estimate of distortion for a random sample of 1K validation CIFAR-10 images. At the model's peak robustness (Epoch 15) distortion is high, and as the model begins to suffer from CO ($\sim$ Epoch 23) the distribution shifts towards lower values until it strongly accumulates at low values when CO has fully taken effect.

Most interestingly about this result is that FW(2) is able to {\bf detect CO through its distortion without needing to fool the model} (it had a success rate of only $16\%$). This points to a strong connection between distortion of FW attacks and robustness which make rigorous below.

\subsection{Multi-step High Distortion Attacks are Inefficient}

Our main tool in analyzing the distortion of \textsc{FW} attacks, and a prime reason \textsc{FW-AT} is more mathematically transparent than \pgdat, is a representation of the \textsc{FW} attack as a convex combination of the \textsc{LMO} iterates. We refer to the steps taken during the optimization as the attack path. Proofs are included in the Appendix.

\begin{prop} \label{prop:FW_adv_pert}
The \textsc{FW} attack with step sizes 
$\gamma_k$
yields the following adversarial perturbation after $K$ steps
%
\begin{equation}
    \delta_K = \epsilon \sum_{l=0}^{K-1} \alpha_l \phi_p(\gradd \loss(x+\delta_l,y))
\end{equation}
where $\alpha_l = \gamma_l \prod_{i=l+1}^{K-1} (1-\gamma_i) \in [0,1]$ are non-decreasing in $l$, and sum to unity.
\end{prop}

Proposition \ref{prop:FW_adv_pert} shows that the \textsc{FW} adversarial perturbation may be expressed as a convex combination of the signed loss gradients for $p=\infty$, and scaled loss gradients for $p\in[1,\infty)$. Using this representation we can deduce connections between the distortion of the attack and the geometric properties of the attack path. 

\begin{theorem}  
\label{thm:FW_distortion}
Consider a $K$ step $\ell_\infty$  \textsc{FW} Attack. 
Let $\cos \beta_{lj}$ be the directional cosine between $\sgn(\gradd \loss(x+\delta_l,y))$ and $\sgn(\gradd \loss(x+\delta_j,y))$. The maximal $\ell_2$ distortion ratio of the adversarial perturbation $\delta_K$ is:
    \begin{equation}
    \label{eq:L2_distortion}
        \frac{\nn \delta_K \nn_2}{\epsilon \sqrt{d} } =
                \sqrt{ 
                        1 - 2 \sum_{l < j} \alpha_l \alpha_j (1-\cos \beta_{lj})
                        }
    \end{equation}
\end{theorem}

We can summarize the spirit of Theorem \ref{thm:FW_distortion} as:

\begin{center}
 { \it Higher distortion is equivalent to lower gradient variation throughout the attack path.}    
\end{center}

Concretely, the accumulation of sign changes between every step of the attack decreases distortion. In the extreme case of maximally distorted attacks, this implies that the attack is at the corner of the $\ell_\infty$ ball which could have had {\bf no changes} to the sign of its gradient between any step on the attack path. Therefore each step was constant and the attack is equivalent to a \fwk{1} attack or \textsc{FGSM}. This is graphically illustrated in Figure \ref{fig:distortion_concept}. Following this logic further, we are able to quantify the distance between different step attacks in terms of the final distortion.

\begin{theorem}  \label{thm:FW_distortion_grad}
Let the same conditions as Theorem \ref{thm:FW_distortion} hold and $K>1$. Assume the maximal $\ell_2$ distortion ratio of the adversarial perturbation satisfies:
\begin{equation*}
    \frac{\nn \delta_K \nn_2}{\epsilon \sqrt{d}} 
        \geq 
            \sqrt{1-\eta}
\end{equation*}
for some $\eta \in (0,1)$. 
Then for all intermediate perturbations $\delta_{k_0}$, with $k_0=1,\dots,K$:
\begin{equation} \label{eq:pert_bound}
    \frac{\|\delta_K - \delta_{k_0}\|_2}{\eps \sqrt{d}}
            \leq C_{k_0,K} \sqrt{\eta}
\end{equation}
where $C_{1,K} = \sqrt{K-1}$ and $C_{k_0, K} = \sqrt{ \frac{ 2 \sum_{l=0}^{K-1}(\alpha^l)^2 \sum_{j=0}^{k_0-1}(\tilde{\alpha}^j)^2}{\alpha^0 \alpha^1} } $ for $k_0 > 1$.

\end{theorem}
%


We can summarize the spirit of Theorem \ref{thm:FW_distortion_grad} as:

\begin{center}
    {\it Multi-step attacks with high distortion are inefficient. }
\end{center}

This suggests that during \textsc{FW-AT} using a large number of steps to approximate the adversarial risk results in diminishing returns once high distortion of the attacks is attained since the final step will be close to the early steps. The other direction is true as well,

\begin{center}
    {\it Models attacked with low distortion perturbations can benefit from training with more steps.}
\end{center}


Intuitively, adversarial attacks with low distortion imply the loss can be maximized at a lower $\ell_2$ radius $\epsilon'\sqrt{d} < \epsilon\sqrt{d}$ than the target radius. These loss landscape irregularities are associated with CO as discussed in Section \ref{sec:CO}. Inspired by these two insights we design a \textsc{FW-AT} algorithm which adapts the number of attack steps used in the optimization based on the distortion of a \fwk{2} attack.

\section{Frank-Wolfe Adversarial Training Algorithm} \label{sec:FW_AT_Adapt}

Pseudocode for the Adaptive Frank-Wolfe adversarial training method (\fwadapt) is provided in Algorithm \ref{alg:FW-AT}. 
The algorithm is graphically depicted in Figure \ref{fig:FW_AT_Adapt_concept} and makes the following modifications to \textsc{PGD-AT}:

\begin{enumerate}[label=(\roman*)]
    \item Adversarial attacks are computed using a \textsc{FW} optimization scheme (Alg. \ref{alg:FW-AA})
    \item For the first $B_m$ batches of each epoch, the distortion of a \fwk{2} attack is monitored. If the mean distortion across these batches is above a threshold $r$ then the number of attack steps $K$ is dropped to $K/2$ for the remainder of the epoch. Alternatively if it is lower than $r$ then $K$ is incremented by $2$.
\end{enumerate}

\begin{algorithm}[H]
    \caption{ Epoch of \fwadapt
    }
    \label{alg:FW-AT}
    \begin{algorithmic}
        \State{\bfseries Input: }{
        Model $f_\theta$, data $\mathcal{D}$, epoch $t$, max batch size $|B|$, max perturbation $\epsilon$, step schedule $\gamma_k$, learning rate $\eta_t$, previous epoch steps $K_0$, max steps $K_1$, max distortion ratio $r$, number of monitoring batches $B_m$.
        }
        \State{\bfseries Result: }{Robust model weights $\theta$, current steps $K$.}
        %
            
            \State{$N_{b},\  d_{m}$ = 0}
            \State{$K=2$} \Comment{Check \fwk{2} Distortion}
            \For{each batch $(x,y) \sim \mathcal{D}$ }
                \State{$\delta$ = FW-Attack$(x,y; K, \gamma_k, p=\infty)$}
                \State{$d_{m} = d_{m} + \|\delta\|_2/(\epsilon \sqrt{d})$} 
                \State{$N_b = N_b + 1$} 
                    \If{$N_b = B_m$} 
                        \If{$d_{m} / B_m > r$}\Comment{Check distortion}
                            \State{$K = \max(1,  \floor*{K_0/2} )$}
                        \Else
                            \State{$K = \min(K_1, K_0+2)$} 
                    \EndIf
                \EndIf
               
            \State $\theta = \theta - \eta_t \frac{1}{|B|} \sum_{i\in B} \grad_\theta \loss(f_\theta(x_i+\delta_i),y_i)$
            
            \EndFor
        
    \end{algorithmic}
\end{algorithm}

Next we analyze the effect of using fewer steps on adversarial training weight updates in the high distortion setting. Our analysis shows that in this setting, AT weight updates are minimally affected and thus our  method does not sacrifice robustness. While loss functions $\loss(f_\theta(x+\delta),y)$ in deep neural networks are non-convex in general, we make the following assumption.
\begin{assumption} \label{assump:Lip}
    The function $\loss$ has $L$-Lipschitz continuous gradients on $B_p(\epsilon)$, i.e., $\nn \grad_\theta \loss(f_\theta(x+u),y) - \grad_\theta \loss(f_\theta(x+v),y) \nn \leq L \nn u - v \nn, \forall u,v \in B_p(\epsilon)$.
\end{assumption}
\noindent Assumption \ref{assump:Lip} is a standard assumption that has been made in several prior works \cite{Sinha:2018, WangFOSC:2019}. Recent works have shown that the loss is semi-smooth in over-parameterized \cite{AllenZhu:2019, Zou:2019, Cao:2020} deep neural networks, and batch normalization provides favorable Lipschitz continuity properties \cite{Santurkar:2018}. This helps justify Assumption \ref{assump:Lip}.

\begin{theorem} \label{thm:weight_update_bound}
Consider a batch update of \textsc{FW-AT} Algorithm \ref{alg:FW-AT} where the high distortion condition of Thm. \ref{thm:FW_distortion_grad} holds on average on examples in a batch $B$, i.e. for some small $\eta\in (0,1)$:
\begin{equation} \label{eq:avg_high_distortion}
    \frac{1}{|B|} \sum_{i \in B} \frac{\nn \delta_i(K) \nn_2}{\epsilon \sqrt{d}} \geq \sqrt{1-\eta}
\end{equation}
where $\delta_i(K)$ denotes the $K$-step \textsc{FW} adversarial perturbation for the $i$-th example in the batch $B$. Let the SGD model weight gradient be given by:
\begin{equation*}
    g(\theta,\delta(K)) = \frac{1}{|B|} \sum_{i \in B} \grad_\theta \loss(f_\theta(x_i+\delta_i(K)),y_i) 
\end{equation*}
Given Assumption \ref{assump:Lip} holds, the model weights SGD update using adversarial perturbations $\delta_K$ and $\delta_{k_0}$ are bounded as:
\begin{equation} \label{eq:g_bound}
    \nn g(\theta,\delta(K)) - g(\theta,\delta(k_0)) \nn_2 \leq L C_{k_0,K} \sqrt{\eta} \cdot \epsilon \sqrt{d}
    .
\end{equation}
\end{theorem}

Bound \eqref{eq:g_bound} asserts that in the high distortion setting, the gradients, and thus the weight updates, obtained by a high-step \textsc{FW} attack are near those of a low-step \textsc{FW} attack.
Therefore it is expected to achieve a similar level of adversarial robustness using the proposed adaptive algorithm. The proof is included in the Appendix.

\subsection{Choosing the Target Distortion Ratio}

To provide intuition for the distortion ratio signal hyperparameter, $r$, we present the following corollary.

\begin{corollary}(\fwk{2} Distortion Check)
\label{cor: fw2 dist}
Let $s_0$ and $s_1$ be the \textsc{LMO}s for the first two steps of a FW adversarial attack. Then if $s_1$ has $k$ sign changes from $s_0$ the maximal distortion of $\delta_1$ is
\begin{equation}
\label{eq: fw2 dist}
    \frac{\|\delta_1\|_2}{\eps \sqrt{d}} = 
        \sqrt{
            1 - \frac{4k}{d}\gamma_1 (1-\gamma_1)
        }
\end{equation}

\end{corollary}

Corollary \ref{cor: fw2 dist} tells us that the distortion of \fwk{2} is a function of the number of sign change ratio from the loss gradient at $x$ and at the FGSM attack. For example Figure \ref{fig: FW-2 Dist} shows that CO models always have more than $30\%$ sign changes between iterates, whereas more robust can have as few as $10\%$ sign changes.

\section{Experimental Results}

We evaluate our models on the CIFAR-10 and CIFAR-100 datasets \cite{CIFAR-datasets} for $\epsilon=8/255$ and $16/255$. All networks were initialized with the weights of a pretrained standard model then fine-tuned for $30$ epochs via SGD optimization. The learning rate was $0.1$ (aside from \textsc{FGSM-GA}) and was then decreased to $0.01$ after $15$ epochs. We record the time to train the full $30$ epochs (aside from \textsc{Free-AT}). For \textsc{FW-Adapt} we choose 15 evenly spaced sign change ratios between 15 and 30\% then set the distortion check based on Corollary \ref{cor: fw2 dist}.

{\bf Baselines.} We compare against multi-step \textsc{PGD(K)-AT} using a step size of $2.5 \epsilon / K$ and $K=2,3,5,7,10$. 
Additionally, we compare against methods which use a single gradient step in their defense. This includes \textsc{FGSM-Rand}, and \textsc{FGSM-Adapt} with a step size of $\epsilon$ and a sweep of checkpoints $c=2, 3, 4, 8$. 
Although it arguably uses multiple steps due to minibatch replays and warm starting of attacks we also include \textsc{Free-AT} in this category, where we sweep the number of minibatch replays $m=2, 3, 4, 8, 12$ since it obtains comparable training times to other single step methods. 
We place \textsc{FGSM-GA} in this category, even though it requires additional gradient information to compute its alignment regularization, since it still attacks with one step. Our \textsc{FW-Adapt} algorithm belongs to a separate category as it aims to efficiently use multiple steps through adaptation, thereby bridging the gap between fixed multi-step and single-step methods.

\textsc{Free-AT} training for $30$ epochs would unfairly increase its training time since the minibatch replay effectively multiplies the number of steps the mode takes. To address this we tuned the number of epochs and minibatches to be near the clean accuracy of competitor methods. Further, \textsc{FGSM-GA} was unable to achieve high accuracy with a learning rate starting at $0.1$ and so had its learning rate set to $0.01$.

{\bf Evaluation Metrics.} Robustness is evaluated using a strong white box attack, \textsc{\pgdk{50}} with step size of $2.5 \epsilon / 50$. To ensure that we detect gradient masking we also evaluate against AutoAttack (AA) \cite{Croce:2020}, a hyperparameter-free attack suite composed of multiple strong white and black box attacks, making it a strong evaluation metric against gradient masking.

\begin{figure}
    \centering
    \begin{subfigure}[b]{0.42 \textwidth}
        \includegraphics[width=0.9\textwidth]{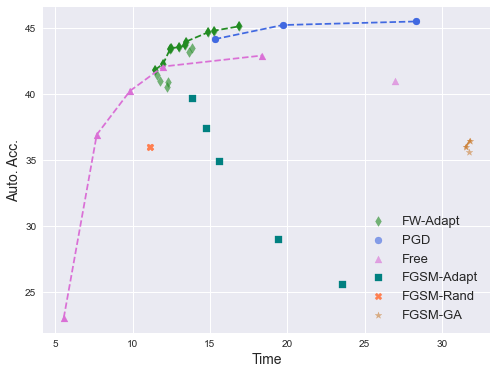}
        \caption{$\epsilon=8/255$}
        \label{fig:aa v time plots c10 eps8}
    \end{subfigure}
    \centering
    \begin{subfigure}[b]{0.42 \textwidth}
        \includegraphics[width=0.9\textwidth]{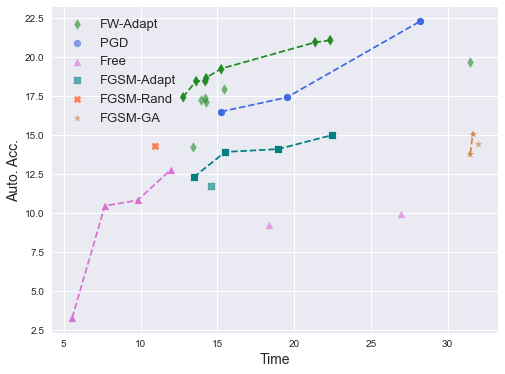}
        \caption{$\epsilon=16/255$}
        \label{fig:aa v time plots c10 eps16}
    \end{subfigure}

\caption{AutoAttack performance vs. training times tradeoffs on CIFAR-10 for various AT methods. Points represent individual parameters of a method and curves are the optimal performance tradeoffs. \textsc{FW-Adapt} achieves superior tradeoffs for similar time complexity methods.}
\label{fig:aa v time plots c10}
\end{figure}

{\bf Results.} Figure \ref{fig:aa v time plots c10} shows the results for parameter sweep on CIFAR-10 of single-step and multi-step methods which trained in less than 35 minutes. Each point represents a different parameter, and the curves show the optimal performance curve which is defined as parameters for which there are none with higher AutoAttack accuracy that trained faster. In general \textsc{FW-Adapt} obtains superior robustness vs. training time tradeoffs and in particular in the more difficult $\epsilon=16/255$ we get substantial improvement over other methods. 

Comparisons at the endpoints of the performance curves for each method are given in Table \ref{tab:main c10 table}. We see \textsc{FW-Adapt} is able to close the gap between single and multi-step methods in terms of robustness without sacrificing speed. In particular we see in the $\epsilon=16/255$ case single step methods struggle in both clean and adversarial accuracy; whereas \textsc{FW-Adapt} is able to achieve similar training times with much higher performance. This suggests that larger attack sizes present a fundamental impediment to using single step methods. Similar performance benefits are observed on the CIFAR-100 dataset in Table \ref{tab:cifar100_table}.

\begin{table}[]
    \centering
    
    \footnotesize
\begin{tabular}{r rrrr}

\toprule
Method $\epsilon=8/255$&   Clean &  \pgdk{50} &  AA & Time (min) \\
    \midrule
    
                  \textsc{\pgdk{10}} &  82.31 &   50.11 & 45.38 & 49.8 \\
                   \textsc{\pgdk{5}} &  82.46 &   49.88 & 45.46 & 28.3 \\
                   \textsc{\pgdk{2}} &  83.42 &   48.50 & 44.12 & 15.3 \\
\midrule
    \textsc{Free-AT} ($m=2$) &  90.10 &   26.74 & 23.01 &        5.5 \\
    \textsc{Free-AT} ($m=12$) &  76.50 &   45.09 & 40.97 &       27.0 \\
   
  \textsc{FGSM-GA} ($\lambda=0.2$) &  77.99 &   41.44 & 36.42 &       31.8 \\
        \textsc{FGSM-Adapt} ($c=2$) &  83.97 &   43.77 & 39.65 & 13.9 \\
               \textsc{FGSM-Rand}  &  78.38 &   40.64 & 35.97 & 11.1 \\
\midrule
      \textsc{FW-Adapt} ($r=0.865$) &  83.31 &   45.81 & 41.80 & 11.5 \\
      \textsc{FW-Adapt} ($r=0.900$) &  82.34 &   49.67 & 45.09 & 16.9 \\

\bottomrule
\toprule
Method $\epsilon=16/255$ &  Clean &  \pgdk{50} &  AA & Time (min) \\ \midrule
                 \textsc{\pgdk{10}} &  63.17 &   31.29 & 22.62 &       49.8 \\
                  \textsc{\pgdk{5}} &  61.72 &   30.58 & 22.27 &       28.2 \\
                  \textsc{\pgdk{2}} &  63.21 &   24.76 & 16.49 &       15.2 \\
 \midrule
                    \textsc{Free-AT} ($m=2$) &  57.69 &    5.00 &  3.27 &        5.5 \\
                     \textsc{Free-AT} ($m=5$) &  35.05 &   15.94 & 12.77 &       12.0 \\
             \textsc{FGSM-GA} ($\lambda=0.5$) &  55.57 &   22.72 & 15.09 &       31.6 \\
      \textsc{FGSM-Adapt} ($c=12$) &  33.18 &   18.82 & 14.99 &       22.5 \\
       \textsc{FGSM-Adapt} ($c=2$) &  40.44 &   17.32 & 12.29 &       13.5 \\
              \textsc{FGSM-Rand} &  56.82 &   22.03 & 14.27 &       11.0 \\
 \midrule
     \textsc{FW-Adapt} ($r=0.830$) &  57.78 &   25.54 & 17.41 &       12.8 \\
     \textsc{FW-Adapt} ($r=0.887$) &  58.46 &   29.57 & 21.06 &       22.4 \\
 
\bottomrule
\end{tabular}

    \caption{Adversarial accuracy computed with \pgdk{50}, AutoAttack (AA) and training time of baseline multi-step \textsc{PGD} (first block), single-step (second block) and \textsc{FW-Adapt} (third block). Results for parameters at end points of performance curves, including \pgdk{10}, for CIFAR-10 dataset. }
    \label{tab:main c10 table}
\end{table}

\begin{table}
\footnotesize
\begin{tabular}{rrrrr}
\toprule
         Method $\epsilon=8/255$ &  Clean &  \pgdk{50} &  AA & Time (min) \\
\midrule
                    \textsc{\pgdk{10}} &  59.07 &   27.37 & 23.10 &       49.8 \\
                    \textsc{\pgdk{2}} &  60.65 &   26.20 & 21.99 &       15.3 \\
\midrule
                   \textsc{Free-AT} ($m=4$) &  60.16 &   23.20 & 19.27 &        9.9 \\
\textsc{FGSM-GA} ($\lambda=0.2$) &  56.53 &   20.02 & 16.15 &       31.5 \\
            \textsc{FGSM-Adapt} ($c=2$) &  49.28 &   20.19 & 15.97 &       13.7 \\
                    \textsc{FGSM-Rand} &  50.20 &   20.63 & 16.47 &       11.0 \\
                   \midrule
                   \textsc{FW-Adapt} ($r=0.830$) &  61.12 &   23.20 & 19.97 &       11.9 \\
           \textsc{FW-Adapt} ($r=0.899$) &  60.60 &   25.41 & 21.64 &       15.1 \\
           
\bottomrule
\toprule
        Method $\epsilon=16/255$ &  Clean &  \pgdk{50} &  AA & Time (min) \\
\midrule
                    \textsc{\pgdk{10}} &  40.49 &   16.46 & 11.30 &       49.8 \\
                    \textsc{\pgdk{5}} &  41.99 &   15.71 & 10.88 &       28.3 \\
                    \textsc{\pgdk{2}} &  33.17 &   10.54 &  7.14 &       15.3 \\
\midrule
                   \textsc{Free-AT} ($m=4$) &  47.06 &    8.83 &  6.34 &        9.8 \\
\textsc{FGSM-GA} ($\lambda=0.2$) &  37.04 &    7.88 &  4.67 &       32.2 \\
            \textsc{FGSM-Adapt} ($c=2$) &   8.38 &    4.85 &  3.55 &       13.5 \\
                    \textsc{FGSM-Rand} &  24.78 &    6.96 &  3.92 &       11.0 \\
                    \midrule
           \textsc{FW-Adapt} ($r=0.830$) &  44.65 &   11.52 &  7.99 &       13.6 \\
           \textsc{FW-Adapt} ($r=0.887$) &  40.91 &   15.33 & 10.47 &       24.0 \\
\bottomrule
\end{tabular}

    \caption{Adversarial accuracy computed with \pgdk{50}, AutoAttack (AA) and training time of baseline multi-step \textsc{PGD} (first block), single-step (second block) and \textsc{FW-Adapt} (third block). Results for select top-performing models, including \pgdk{10}, for CIFAR-100 dataset.}
    \label{tab:cifar100_table}
\end{table}

\section{Limitations}
Our work focuses on obtaining a deeper understanding of the theory behind \textsc{FW-AT} and establishing whether an adaptive version of \textsc{FW-AT}, \fwadapt, can offer superior robustness / training time tradeoffs compared to single-step and multi-step AT variants. We showed indeed such superior tradeoffs exist. 
Future work may focus on developing alternative adaptation strategies and criteria.



\section{Conclusion}
Adversarial training (AT) provide robustness against $\ell_p$-norm adversarial perturbations computed using projected gradient descent (\textsc{PGD}). Through the use of Frank-Wolfe (FW) optimization for the inner maximization an interesting phenomenon occurs: \textsc{FW} attacks against robust models result in higher $\ell_2$ distortions than standard ones despite achieving nearly the same $\ell_\infty$ distortion. We derive a theoretical connection between loss gradient alignment along the attack path and the distortion of \textsc{FW} attacks which explains this phenomenon. We provide theoretical and empirical evidence that this distortion can signal catastrophic overfitting in single-step fast AT models. Inspired by this connection, we propose an adaptive Frank-Wolfe adversarial training (\textsc{FW-AT-Adapt}) algorithm that achieves robustness above single-step baselines while maintaining competitive training times particularly in the strong $\ell_\infty$ attack regime. This work begins to close the gap between robustness training time trade-offs of single-step and multi-step methods and hope it will inspire future research on the connection between Frank-Wolfe optimization and adversarial robustness.


\section*{Societal Impact Statement}
As DNNs are increasingly being deployed for safety-critical applications, such as healthcare, autonomous driving, and biometrics,  robustness against adversarial attacks is a rising concern. Addressing this is critical to gain public trust and avoid denial of opportunity. One of the most popular and effective defenses is adversarial training (AT). However, the popular multi-step PGD optimization approach used in AT cannot be easily analyzed to obtain insights into what type of regularization AT induces, and it also requires multiple steps in the inner maximization leading to slow training. To reduce training time, single-step approaches have been proposed, but are prone to catastrophic overfitting, leading to a false sense of robustness. This can have severe consequences in security applications. Our work improves the understanding of AT via the lens of FW optimization and provides simple methods for efficient training of robust models without compromising robustness.

\section*{Acknowledgements}
Research was sponsored by the United States Air Force Research Laboratory and the United States Air Force Artificial Intelligence Accelerator and was accomplished under Cooperative Agreement Number FA8750-19-2-1000. The views and conclusions contained in this document are those of the authors and should not be interpreted as representing the official policies, either expressed or implied, of the United States Air Force or the U.S. Government. The U.S. Government is authorized to reproduce and distribute reprints for Government purposes notwithstanding any copyright notation herein.

{\small
\bibliographystyle{abbrv}
\bibliography{refs}
}

\clearpage
\begin{appendices}

\section{Results on Additional Datasets}

\begin{figure}[h]
    \centering
    \begin{subfigure}[b]{0.4\textwidth}
         \centering
         \includegraphics[width=\textwidth]{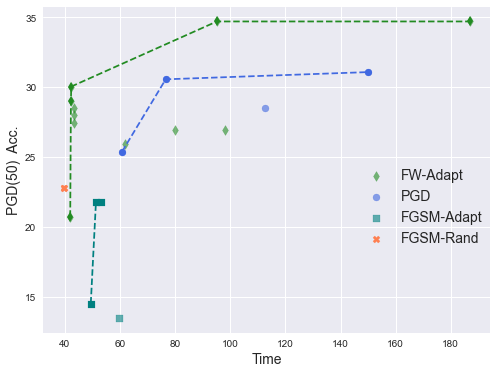}
         \caption{ISIC-2018}
     \end{subfigure}
     \hfill
     \begin{subfigure}[b]{0.4\textwidth}
         \centering
         \includegraphics[width=\textwidth]{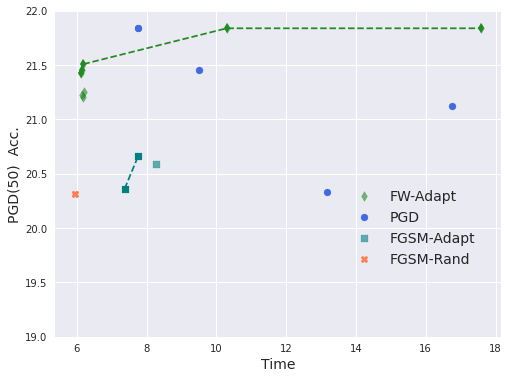}
         \caption{ImageWoof}
     \end{subfigure}
    \caption{Adversarial accuracy against $\epsilon=8/255$ attacks with \pgdk{50}. Dashed line spans optimal parameters. (PGD steps 2,3,5,7)
    }
    \label{fig:extra data}
\end{figure}

We further validate our method on two higher resolution datasets. The first is a dataset of 7 types of skin lesions at a resolution of $224 \times 224$, ISIC-2018 \cite{isic2018}, and the second is a subset of ImageNet where the task is to classify 10 breeds of dogs resized to $64\times64$, ImageWoof. Both datasets have $\sim$10K/1K train/val images. Due to computation constraints, we show only results against \pgdk{50} attack. Figure \ref{fig:extra data} shows the results of these experiments. 

For the ISIC-2018 dataset we see an even stronger trend than with CIFAR-10/100. Here \fwadapt almost uniformly outperforms competing methods with respect to optimal tradeoffs. A similar trend to CIFAR-10/100 holds for ImageWoof; however, the results are much less pronounced. In particular PGD struggled at higher steps. We suspect the lower performance is due to the lower resolution increasing the difficulty of differentiating dog breeds which share many semantically similar features, although we note \fwadapt seems to maintain performance across parameters.

\section{Effects of Minimum Distortion Ratio Bound on FW-Adapt}

\begin{figure}[]
    \centering
    \begin{subfigure}[b]{0.4 \textwidth}
        \includegraphics[width=0.99\textwidth]{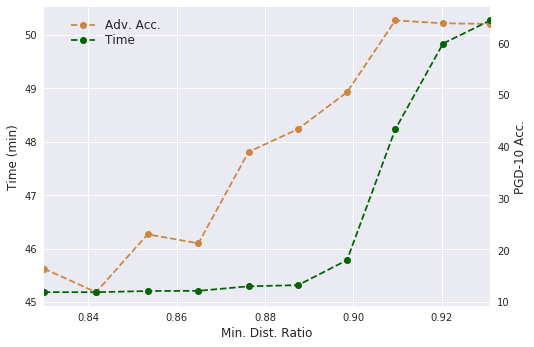}
        \caption{$\epsilon=8/255$}
    \end{subfigure}
    \hfill
    \begin{subfigure}[b]{0.4 \textwidth}
        \includegraphics[width=0.99\textwidth]{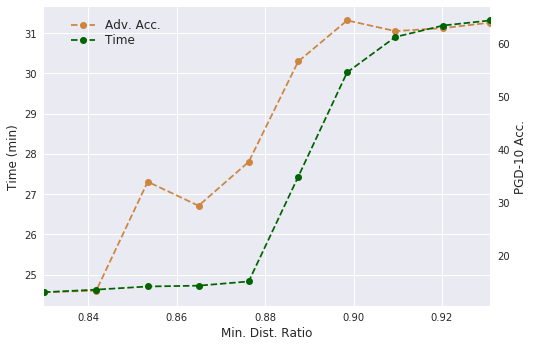}
        \caption{$\epsilon=16/255$}
    \end{subfigure}
    \caption{Average Training times and adversarial accuracy against \pgdk{10} as a function of minimum distortion ratio over five independent runs.}
    \label{fig:sup fwadapt mdr train time}
\end{figure}

To better understand the impact of the minimum distortion ratio, $r$, on \textsc{FW-Adapt} we run the same distortion ratios as in our main set of experiments over $5$ independent runs, and analyzed various performance and training metrics. In Figure \ref{fig:sup fwadapt mdr train time} plots both training time and adversarial accuracy as a function of the minimum distortion ratio bound, $r$. Adversarial accuracy is computed against a \pgdk{10} attack with step size $2.5 \epsilon / 10$. Both time and adversarial accuracy are reported as the mean of $5$ independent training runs. We see for both $\epsilon=8/255$ and $16/255$ the adversarial accuracy increases with training time. In both cases, there seems to be an optimal $r$ in terms of training time vs robustness tradeoffs, around $0.9$ and $0.88$ for $\epsilon=8/255$ and $16/255$ respectively.

\begin{figure}[]
    \centering
    \begin{subfigure}[b]{0.45 \textwidth}
        \includegraphics[width=0.99\textwidth]{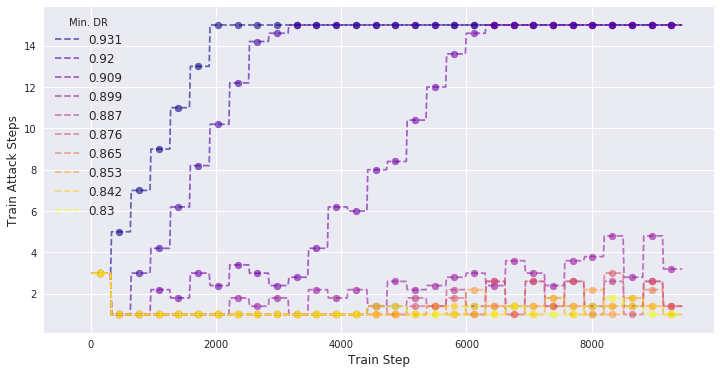}
        \caption{$\epsilon=8/255$}
    \end{subfigure}
    \hfill
    \begin{subfigure}[b]{0.45 \textwidth}
        \includegraphics[width=0.99\textwidth]{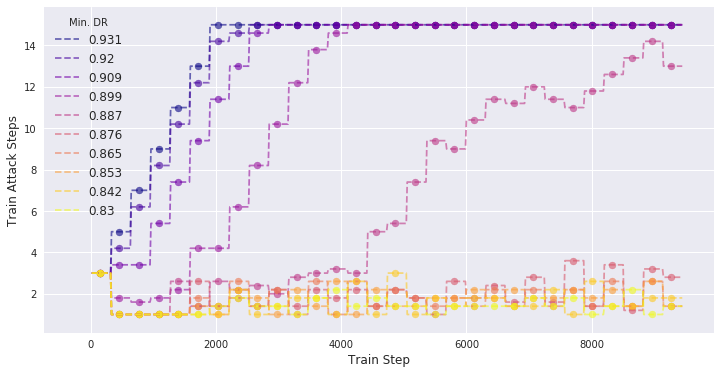}
        \caption{$\epsilon=16/255$}
    \end{subfigure}
    \caption{Number of attack steps during training for varying minimum distortion ratios. Plot ignores the first batch which is always done with 2 steps. Results averaged over 5 runs.}
    \label{fig:sup fwadapt steps train}
\end{figure}

In Figure \ref{fig:sup fwadapt steps train} we show how the number of steps used by \textsc{FW-Adapt} evolves during training for our different values of the minimum distortion ratio bound $r$. We do not consider the first batch as this is always a two-step attack to monitor the distortion. Steps are averaged across $5$ independent runs.

Higher values of $r$ result in a linear increase towards the maximum number of steps, $15$, and very low values of $r$ result in primarily, although importantly not exclusively, single steps of attacks during training. As expected, the optimal value of $r$ based on Figure \ref{fig:sup fwadapt mdr train time} corresponds to training strategies which used a small number of steps initially and then modestly increase during training. 

\begin{figure}[]
    \centering
    \begin{subfigure}[b]{0.4 \textwidth}
        \includegraphics[width=0.99\textwidth]{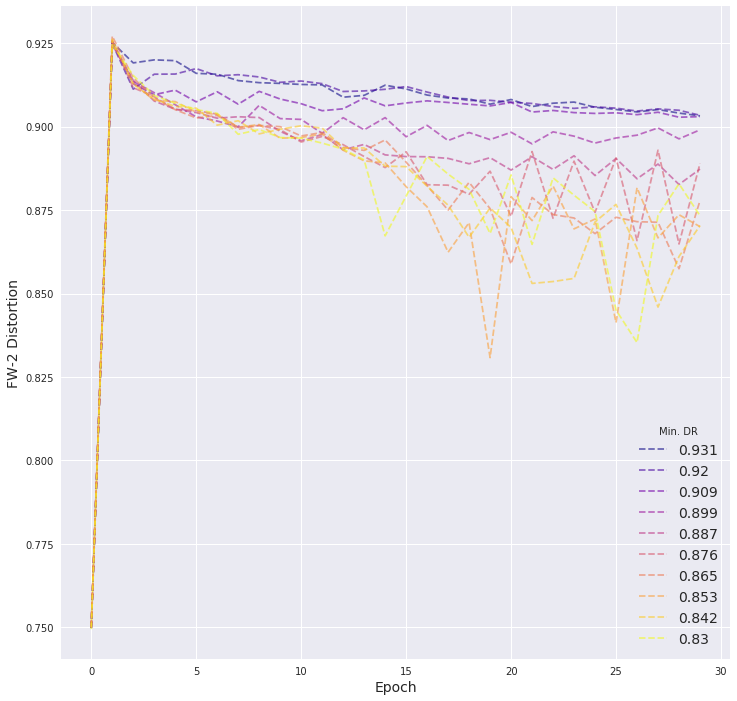}
        \caption{$\epsilon=8/255$}
    \end{subfigure}
    \hfill
    \begin{subfigure}[b]{0.4 \textwidth}
        \includegraphics[width=0.99\textwidth]{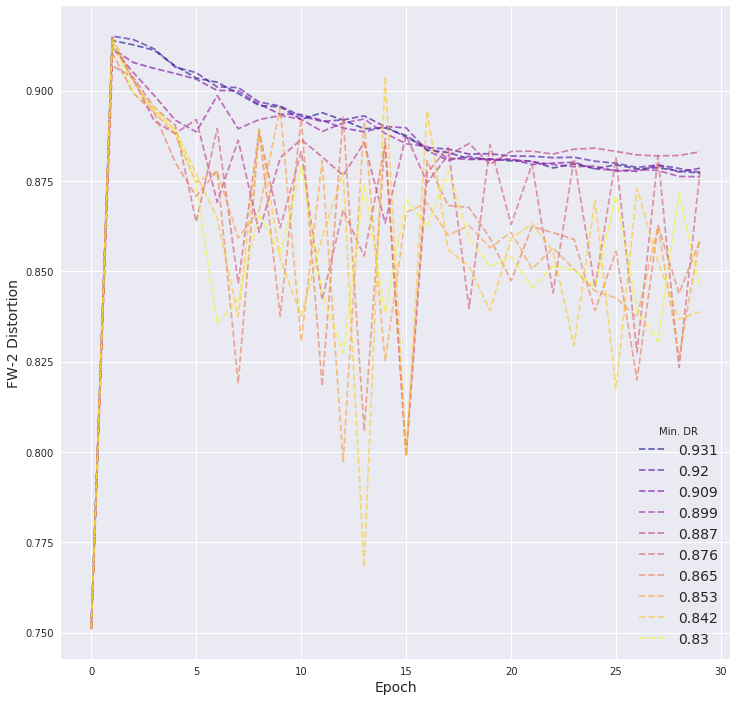}
        \caption{$\epsilon=16/255$}
    \end{subfigure}
    \caption{Average distortion check value during training for varying minimum distortion ratios.Results averaged over 5 runs.}
    \label{fig:sup fwadapt dist train}
\end{figure}

Figure \ref{fig:sup fwadapt dist train} is in a sense dual to Figure \ref{fig:sup fwadapt steps train} in that we plot the value of \textsc{\fwk{2}} distortion used in the adaptive step check. Again, we averaged the values over five independent runs.

The high values of $r$ which quickly increased their training steps have a smooth gradual decay in their distortion check; whereas, lower values had much more variation in their checks. The overall trend of decaying distortion is interesting and reinforces the fact that as AT progresses, stronger multi-step attacks are needed to more effectively increase the loss, but early in training such steps are not necessary. \textsc{FW-Adapt} is able to capitalize on this to achieve faster training times. In future work, we hope to better understand the decaying trend of distortion, and perhaps develop more sophisticated adaptive criterion and step modifications to further improve performance.

\section{FW-AT is As Good As PGD-AT}
\begin{figure}[t]
    \centering
    \begin{subfigure}[b]{0.4 \textwidth}
        \includegraphics[width=0.99\textwidth]{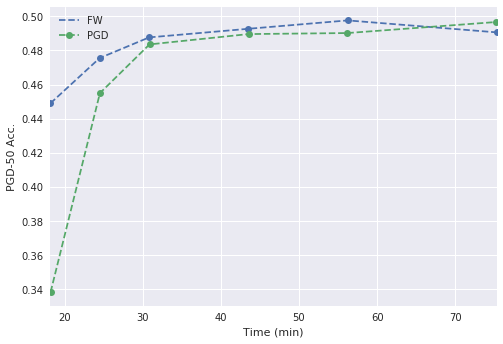}
        \caption{$\epsilon=8/255$}
    \end{subfigure}
    \hfill
    \begin{subfigure}[b]{0.4 \textwidth}
        \includegraphics[width=0.99\textwidth]{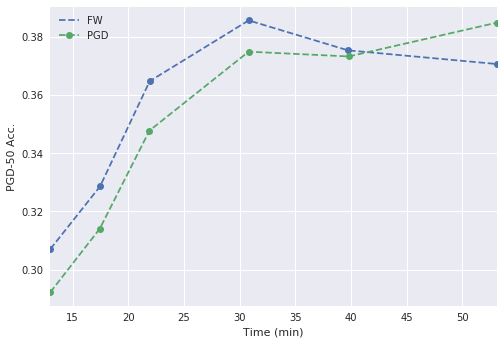}
        \caption{$\epsilon=16/255$}
    \end{subfigure}
    \caption{Accuracy against \pgdk{5}0 attacks on CIFAR-10 validation images for FW-AT and PGD-AT at various $\epsilon$ values.}
    \label{fig:sup fw-pgd train time}
\end{figure}

Although we focus on the novel \textsc{FW-Adapt} algorithm here, we note that using FW optimization (Algorithm 1) in place of PGD with no other alterations performs as well as PGD in terms of robustness and training times. Figure \ref{fig:sup fw-pgd train time} shows the training times and accuracy against \pgdk{5}0 attacks for models trained with PGD-K-AT and FW-K-AT for $K\in \{1,2,3,5,7,10\}$. The training parameters are the same as those above except accuracy and training time are averaged over three independent runs and we train for 40 epochs. We see that FW-AT performs comparably to PGD-AT. We hope this will encourage further study of FW for deep learning and AT.

\section{Distortion and Gradient Alignment as Catastrophic Overfitting Signals}

\begin{figure}[t]
    \centering
    \begin{subfigure}[b]{0.45 \textwidth}
        \includegraphics[width=0.99\textwidth]{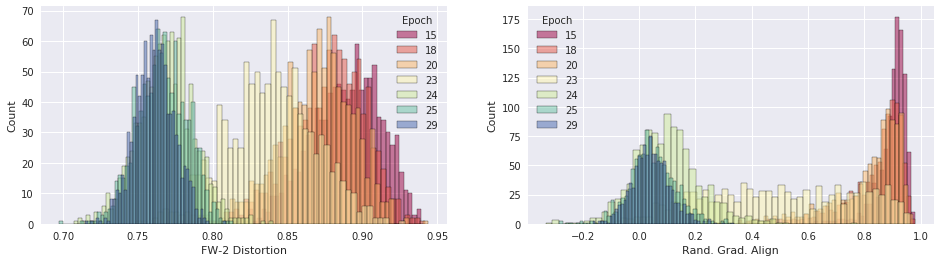}
    \end{subfigure}
    \hfill
    \begin{subfigure}[b]{0.45 \textwidth}
        \includegraphics[width=0.99\textwidth]{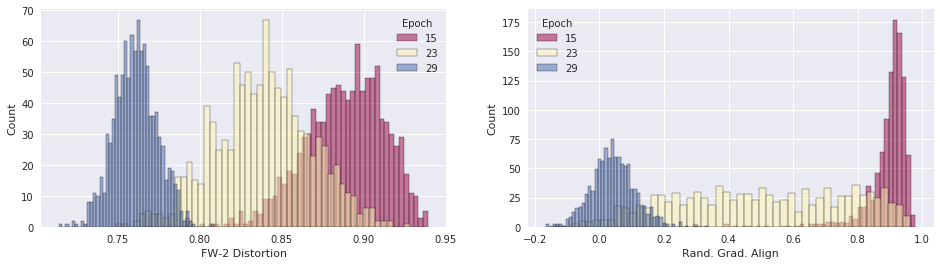}
    \end{subfigure}
    \caption{Distribution of \fwk{2} Distortion and Gradient direction at the point and and random direction (Grad. Align signal) for 1024 randomly sampled CIFAR-10 validation images during FGSM-AT training where catastrophic overfitting occurs as in Figure 4. (Top) The transition across 7 epochs into CO, (Bottom) focusing on epochs clearly before, during, and clearly after CO occurs.}
    \label{fig:sup dist plot}
\end{figure}

The basis of \textsc{FW-Adapt} is that a small number of batches going through low-compute adversarial training (\textsc{\fwk{2}}) can provide a strong signal as to how many steps are needed for the rest of the epoch. As a particular example of this we showed empirically that the distortion of \textsc{\fwk{2}} attacks is a strong signal of catastrophic overfitting (CO), the phenomena where a model is trained with a single step attack and is achieving high accuracy against strong multi-step attacks, but then suddenly loses robustness against strong attacks while still being robust to the single step attack. In \cite{Andriushchenko:2020} authors note that gradient misalignment is also strongly associated with CO and they use it to regularize single step methods (\textsc{FGSM-GA}). 

Here we compare \fwk{2} distortion and gradient alignment (GA) as signals for CO. In figure \ref{fig:sup dist plot} shows the gradual (top) and overall transition of model into CO. Both the distributions of GA and \fwk{2} distortion are able to distinguish the CO model from the non-CO model. During the transition we see the GA score distribution becomes more diffuse during the transition
; whereas, the \fwk{2} distortion has a more gradual peak shift during transition. 

Interestingly, the GA signal may be slightly clearer than the \fwk{2} distortion for CO detection. Although, as we see above, using the GA as a regularizer for single step methods is not able to achieve the same level of robustness as multi-step methods. This suggest that there is more to the gap between single and multi-step methods than merely fixing CO.
Building upon the theoretical foundation for exactly what is missed by single step methods is an interesting direction of further research.

\section{Proofs}

\subsection{Proof of Proposition 1}


\begin{proof}
The \textsc{LMO} solution is given by $\bar{\delta}_k = \epsilon \ \phi_p(\gradd \loss(x+\delta_k,y))$ and the update becomes
\begin{align*}
    \delta_{k+1} &= \delta_k + \gamma_k (\bar{\delta}_{k}-\delta_k) \\
     &= (1-\gamma_k) \delta_k + \gamma_k \ \epsilon \ \phi_p(\grad_\delta \loss(x+\delta_k,y))
\end{align*}
Using induction on this relation yields after $K$ steps:
\begin{align}
    \delta_K &= \delta_0 \prod_{l=0}^{K-1} (1-\gamma_l) \nonumber \\
        &+ \epsilon \sum_{l=0}^{K-1} \gamma_l \prod_{i=l+1}^{K-1} (1-\gamma_i) \phi_p(\gradd \loss(x+\delta_k,y))  \label{eq:deltaK}
\end{align}
where $\delta_0$ is the initial point which affects both terms in (\ref{eq:deltaK}) and $\gamma_k = c/(c+k)$ for $k\geq 0$. Since $\gamma_0=1$, the first term vanishes and (\ref{eq:deltaK}) simplifies to
\begin{equation} \label{eq:deltaK2}
    \delta_K = \epsilon \sum_{l=0}^{K-1} \alpha_l \phi_p(\gradd \loss(x+\delta_l,y))
\end{equation}
where the coefficients are
\begin{equation} \label{eq:alpha}
    \alpha_l = \gamma_l \prod_{i=l+1}^{K-1} (1-\gamma_i)
\end{equation}
Since $\gamma_l \in [0,1]$, it follows that $\alpha_l \in [0,1]$. Induction on (\ref{eq:alpha}) yields that $\sum_{l=0}^{K-1} \alpha_l = 1$. Furthermore, $\alpha_l \leq \alpha_{l+1}$ follows from:
\begin{align*}
    \alpha_l &\leq \alpha_{l+1} \\
    \Leftrightarrow \gamma_l \ (1-\gamma_{l+1}) &\leq \gamma_{l+1} \\
    \Leftrightarrow	\frac{c}{c+l} \Big(1-\frac{c}{c+l+1}\Big) &\leq \frac{c}{c+l+1} \\
    \Leftrightarrow	\frac{l+1}{c+l} &\leq 1 \\
    \Leftrightarrow	1 &\leq c
\end{align*}
Thus, the sequence $\alpha_l$ is non-decreasing in $l$. Since the coefficients sum to unity, (\ref{eq:deltaK2}) is in the convex hull of the generated \textsc{LMO} sequence $\{\phi_p(\gradd \loss(x+\delta_l)):l=0,\dots,K-1\}$.
\end{proof}

\subsection{Proof of Theorem 1}

\begin{proof}
From Proposition 1, we obtain the following decomposition of the adversarial perturbation:
\begin{equation*}
    \delta_K = \epsilon \sum_{l=0}^{K-1} \alpha_l \sgn(\gradd \loss(x+\delta_l,y))
\end{equation*}
To bound the magnitude of the adversarial perturbation, we have
\begin{equation*}
    \nn \delta_K \nn_2 = \sqrt{ \nn \delta_K \nn_2^2 } = \epsilon \sqrt{ \Big\nn \sum_l \alpha_l s_l \Big\nn_2^2}
\end{equation*}
where we use the shorthand notation $s_l = \sgn(\gradd \loss(x+\delta_l,y))$. The squared $\ell_2$ norm in the above is bounded as:
\begin{align*}
    \Big\nn &\sum_l \alpha_l s_l \Big\nn_2^2 = \sum_l \sum_j \alpha_l \alpha_j \left<s_l, s_j \right> \\
        &= \sum_l (\alpha_l)^2 \nn s_l \nn_2^2 + \sum_{l\neq j} \alpha_l \alpha_j \nn s_l\nn_2  \nn s_j \nn_2 \cos \beta_{lj} \\
        &= d \Big( \sum_l (\alpha_l)^2 + \sum_{l\neq j} \alpha_l \alpha_j \cos \beta_{lj} \Big) \\
        &= d \Big( \sum_l (\alpha_l)^2 + \sum_{l\neq j} \alpha_l \alpha_j - \sum_{l\neq j} \alpha_l \alpha_j (1-\cos \beta_{lj}) \Big) \\
        &= d \Big( 1 - \sum_{l\neq j} \alpha_l \alpha_j (1-\cos \beta_{lj}) \Big) \\
        &= d \Big( 1 - 2 \sum_{l < j} \alpha_l \alpha_j (1-\cos \beta_{lj}) \Big)
\end{align*}
where we used $\nn s_l \nn_2 = \sqrt{d}$ and from Proposition 1 $(\sum_l \alpha_l)^2 = 1$. The final step follows from symmetry. This concludes the proof.
\end{proof}

\subsection{Proof of Theorem 2}


\begin{proof}
From Theorem 1 and the lower bound on the distortion, it follows that:
\begin{equation} \label{eq:ineq}
    \sum_{l<j} \alpha_l \alpha_j (1-\cos\beta_{lj}) \leq \eta/2
\end{equation}
Letting $s_i=\text{sgn}(\grad \loss(x+\delta_i,y))$ and expanding the squared difference of signed gradients:
\begin{align}
    \nn s_l-s_j \nn_2^2 &= \nn s_l \nn_2^2 + \nn s_j \nn_2^2 - 2\left<s_j,s_l\right> \nonumber \\
        &= \nn s_l \nn_2^2 + \nn s_j \nn_2^2 - 2 \nn s_j \nn_2 \nn s_l \nn_2 \cos \beta_{lj} \nonumber \\
        &= d + d - 2d \cos \beta_{lj} \nonumber \\
        &= 2d (1-\cos \beta_{lj}) \label{eq:s_cos}
\end{align}
Using (\ref{eq:s_cos}) into (\ref{eq:ineq}),
\begin{equation} \label{eq:ineq2}
    \sum_{l<j} \alpha_l \alpha_j \nn s_l-s_j \nn_2^2 \leq \eta d
\end{equation}

For the FGSM deviation bound, i.e., $k_0=1$, we have by the triangle inequality:
\begin{align}
    \|\delta_K-\epsilon \sgn(\grad_x \loss(x,y))\|_2 &= \| \epsilon \sum_{l=0}^{K-1} \alpha_l s_l - \epsilon s_0 \|_2 \nonumber \\
        &= \| \epsilon \sum_l \alpha_l s_l - \sum_l \alpha_l \epsilon s_0 \|_2 \nonumber \\
        &= \epsilon \| \sum_l \alpha_l (s_l - s_0) \|_2 \nonumber \\
        &\leq \epsilon \sum_{l>0} \alpha_l \| s_l-s_0 \|_2 \label{eq:delta}
\end{align}
Using Cauchy-Schwarz inequality, we obtain:
\begin{align}
    \sum_{l>0} \alpha_l \| s_l-s_0 \|_2 &\leq \sqrt{K-1} \sqrt{\sum_{l>0} (\alpha_l)^2 \|s_l-s_0\|_2^2} \nonumber \\
        &\leq \sqrt{K-1} \sqrt{\sum_{l<j} (\alpha_l)^2 \|s_l-s_j\|_2^2} \nonumber \\
        &\leq \sqrt{K-1} \sqrt{\sum_{l<j} \alpha_l \alpha_j \|s_l-s_j\|_2^2} \nonumber \\
        &\leq \sqrt{K-1} \cdot \sqrt{\eta d} \label{eq:cs}
\end{align}
where we used the non-decreasing property of the sequence $\{\alpha_l\}_l$ and the bound (\ref{eq:ineq2}). This concludes the first part.

Given $1\leq k_0 \leq K$, we have via using Proposition 1 twice:
\begin{align}
    \delta_K - \delta_{k_0} &= \epsilon \sum_{l=0}^{K-1} \alpha_l s_l - \delta_{k_0} \nonumber \\
        &= \epsilon \sum_{l=0}^{K-1} \alpha_l s_l - \sum_l \alpha_l \delta_{k_0} \nonumber \\
        &= \epsilon \sum_{l=0}^{K-1} \alpha_l (s_l - \delta_{k_0}/\epsilon) \nonumber \\
        &= \epsilon \sum_{l=0}^{K-1} \alpha_l (s_l - \sum_{j=0}^{k_0-1} \tilde{\alpha}_j s_j ) \nonumber \\
        &= \epsilon \sum_{l=0}^{K-1} \alpha_l \sum_{j=0}^{k_0-1} \tilde{\alpha}_j ( s_l - s_j ) \nonumber \\
        &= \epsilon \sum_{l=0}^{K-1} \sum_{j=0}^{k_0-1} \alpha_l \tilde{\alpha}_j ( s_l - s_j ) \label{eq:dk_diff}
\end{align}
where $\alpha_l = \gamma_l \prod_{i=l+1}^{K-1}(1-\gamma_i),0\leq l\leq K-1$ and $\tilde{\alpha}_j=\gamma_j \prod_{i=l+1}^{k_0-1}(1-\gamma_i), 0\leq j\leq k_0-1$.

Taking the $\ell_2$ norm of both sides of (\ref{eq:dk_diff}) and using the triangle inequality, we obtain:
\begin{equation*}
    \parallel \delta_K - \delta_{k_0} \parallel_2 \leq \epsilon \sum_{l=0}^{K-1} \sum_{j=0}^{k_0-1} \alpha_l \tilde{\alpha}_j \nn s_l - s_j \nn_2
\end{equation*}
Using the Cauchy-Schwarz inequality yields:
\begin{align*}
    &\sum_{l=0}^{K-1} \sum_{j=0}^{k_0-1} \alpha_l \tilde{\alpha}_j \nn s_l - s_j \nn_2 \\
    &\leq \sqrt{\sum_{l=0}^{K-1} \sum_{j=0}^{k_0-1} (\alpha_l \tilde{\alpha}_j)^2} \sqrt{\sum_{l=0}^{K-1} \sum_{j=0}^{k_0-1} \nn s_l - s_j \nn_2^2}  \\
    &\leq \sqrt{\sum_{l=0}^{K-1} (\alpha_l)^2 \sum_{j=0}^{k_0-1} (\tilde{\alpha}_j)^2} \sqrt{\sum_{l=0}^{K-1} \sum_{j=0}^{K-1} \nn s_l - s_j \nn_2^2}  \\
    &\leq \sqrt{\sum_{l=0}^{K-1} (\alpha_l)^2 \sum_{j=0}^{k_0-1} (\tilde{\alpha}_j)^2} \sqrt{ \frac{2 \eta d}{\min_{l<j}\{\alpha_l \alpha_j\}} }  \\
    &= \sqrt{ \frac{2 \sum_{l=0}^{K-1} (\alpha_l)^2 \sum_{j=0}^{k_0-1} (\tilde{\alpha}_j)^2}{\alpha_0 \alpha_1}} \sqrt{\eta d}
\end{align*}
where we used (\ref{eq:ineq2}) and the nondecreasing sequence $\{\alpha_l\}$ implies $\min_{l<j}\{\alpha_l \alpha_j\} = \alpha_0 \alpha_1$. This concludes the proof of the second part.
\end{proof}

\subsection{Proof of Theorem 3}
\begin{proof}
Using the triangle inequality and the $L$-Lipschitz continuous loss gradient assumption:
\begin{align}
    &\nn g(\theta,\delta(K)) - g(\theta,\delta(k_0)) \nn_2 \nonumber \\
    &=     \Big\nn \frac{1}{|B|} \sum_{i \in B} (\grad_\theta \loss(f_\theta(x_i+\delta_i(K)),y_i) \nonumber \\
    &\qquad - \grad_\theta \loss(f_\theta(x_i+\delta_i(k_0)),y_i)) \Big\nn_2 \nonumber \\ 
    &\leq \frac{1}{|B|} \sum_{i \in B} \nn \grad_\theta \loss(f_\theta(x_i+\delta_i(K)),y_i) \nonumber \\
    &\qquad - \grad_\theta \loss(f_\theta(x_i+\delta_i(k_0)),y_i)) \nn_2 \nonumber \\
    &\leq \frac{L}{|B|} \sum_{i \in B} \nn \delta_i(K)-\delta_i(k_0) \nn_2 \label{eq:Lb}
\end{align}
The average distortion condition yields via Proposition 1 (with the superscript $(i)$ denoting the $i$-th example variables):
\begin{equation*}
    \frac{1}{|B|}\sum_{i\in B} \sqrt{1 - 2\sum_{l<j} \alpha_l \alpha_j (1-\cos \beta_{lj}^{(i)})} \geq \sqrt{1-\eta}
\end{equation*}
Using Jensen's inequality (and the concavity of the square root function) further yields after some algebra:
\begin{equation*}
    \frac{1}{|B|} \sum_{i\in B} \sum_{l<j} \alpha_{l} \alpha_{j} (1-\cos \beta_{lj}^{(i)}) \leq  \frac{\eta}{2}
\end{equation*}
Borrowing the relation (\ref{eq:s_cos}) from the proof of Theorem 2, we further obtain:
\begin{equation} \label{eq:ub1}
    \frac{1}{|B|} \sum_{i\in B} \sum_{l<j} \alpha_{l} \alpha_{j} \nn s_l^{(i)}-s_j^{(i)} \nn_2^2  \leq  \eta d
\end{equation}
Using the relation (\ref{eq:dk_diff}), it follows:
\begin{align}
    &\frac{1}{|B|}\sum_{i\in B} \nn \delta_i(K)-\delta_i(k_0) \nn_2 \nonumber \\
    &\stackrel{\mathclap{\normalfont\mbox{(a)}}}{\leq} \frac{1}{|B|} \sum_{i\in B} \epsilon \sum_{l=0}^{K-1} \sum_{j=0}^{k_0-1} \alpha_l \tilde{\alpha}_j \nn s_l^{(i)}-s_j^{(i)} \nn_2 \nonumber \\
    &\stackrel{\mathclap{\normalfont\mbox{(b)}}}{\leq} \epsilon \sqrt{\sum_{l=0}^{K-1} \alpha_l^2 \sum_{j=0}^{k_0-1} \tilde{\alpha}_j^2 } \cdot \frac{1}{|B|} \sum_{i\in B}  \sqrt{ \sum_{l=0}^{K-1} \sum_{j=0}^{k_0-1} \nn s_l^{(i)}-s_j^{(i)} \nn_2} \nonumber \\
    &\stackrel{\mathclap{\normalfont\mbox{(c)}}}{\leq} \epsilon \sqrt{\sum_{l=0}^{K-1} \alpha_l^2 \sum_{j=0}^{k_0-1} \tilde{\alpha}_j^2 } \cdot \sqrt{ \frac{1}{|B|} \sum_{i\in B} \sum_{l=0}^{K-1} \sum_{j=0}^{k_0-1} \nn s_l^{(i)}-s_j^{(i)} \nn_2} \label{eq:ub2}
\end{align}
where we used (a) triangle inequality, (b) Cauchy-Schwarz, and (c) Jensen's inequality.

From (\ref{eq:ub1}), it follows that:
\begin{equation} \label{eq:ub3}
    \frac{1}{|B|} \sum_{i \in B} \sum_{l=0}^{K-1} \sum_{j=0}^{k_0-1} \nn s_l^{(i)}-s_j^{(i)}\nn_2^2 \leq \frac{2 \eta d}{\alpha_0 \alpha_1}
\end{equation}
Combining (\ref{eq:ub3}) with (\ref{eq:ub2}) yields:
\begin{equation*}
    \frac{1}{|B|} \sum_{i\in B} \nn \delta_i(\theta,K)-\delta_i(\theta,k_0) \nn_2
    \leq \epsilon \sqrt{d} \sqrt{\eta} C_{k_0,K}
\end{equation*}
where $C_{k_0,K}=\sqrt{\frac{2 \sum_{l=0}^{K-1} \alpha_l^2 \sum_{j=0}^{k_0-1} \tilde{\alpha}_j^2}{\alpha_0\alpha_1}}$. Using this bound in (\ref{eq:Lb}) concludes the proof.
\end{proof}

\subsection{Convergence Analysis}

Loss functions $\loss(x+\delta,y)$ in deep neural networks are nonconvex in general. For a targeted attack that aims to fool the classifier to predict a specific label, without loss of generality, we seek to minimize the loss $f(\delta)=\loss(x+\delta,y')$ over a $\ell_p$ constraint set. The untargeted case follows similarly. \footnote{For untargeted attacks, $\min_{\delta \in B(\epsilon)} -\loss(x+\delta,y)$ is considered and the FW gap becomes (\ref{eq:FW_gap}).} For general nonconvex constrained optimization, the Frank-Wolfe gap given by \cite{FW:1956}:
\begin{equation}  \label{eq:FW_gap}
    G(\delta_k) 
        = 
            \max_{\delta \in B_p(\epsilon)} 
                \left<\delta-\delta_k, \grad_\delta \loss(x+\delta_k,y) \right>
\end{equation}
is nonnegative in general and zero at stationary points. The convergence of FW on non-convex functions has been studied in \cite{LacosteJulien:2016} and recently improved for strongly convex constraints in \cite{RectorBrooks:2019}.

%
\begin{assumption} \label{assump-Lip}
    The function $f$ has $L$-Lipschitz continuous gradients on $B_p(\epsilon)$, i.e., $\nn \grad f(u) - \grad f(v) \nn \leq L \nn u - v \nn, \forall u,v \in B_p(\epsilon)$.
\end{assumption}
\noindent Assumption \ref{assump-Lip} is a standard assumption for the nonconvex setting and has been made in several works \cite{LacosteJulien:2016, Chen:2020}. A recent study \cite{Santurkar:2018} shows that the batch normalization layer used in modern neural networks makes the loss much smoother. Furthermore, the process of adversarial training smooths the loss landscape in comparison to standard models significantly \cite{md2019:cure, qin:2019}.

Given Assumption \ref{assump-Lip} and the compactness of the constraint sets, all limit points of FW are stationary points \cite{Bertsekas}. The convergence rate of FW to a stationary point for optimization over arbitrary convex sets was first shown in \cite{LacosteJulien:2016} given by

\begin{equation*}
    \min_{1 \leq s \leq t} G(\delta_s) \leq  \frac{\max\{2 h_0, L \ \diam(B)\}}{\sqrt{t+1}}
\end{equation*}

where $h_0 = f(\delta_0)-\min_{\delta \in B(\epsilon)} f(\delta)$ is the initial global suboptimality. It follows that larger $\epsilon$ imply a larger diameter and more iterations may be needed to converge \footnote{The diameter of $\ell_2$ ball is $2\epsilon$ and for the $\ell_\infty$ ball $2\epsilon \sqrt{d}$.}. This result implies that an approximate stationary point can be found with gap less than $\epsilon_0$ in at most $O(1/\epsilon_0^2)$ iterations. Theorem 4 in \cite{RectorBrooks:2019} shows that for smooth non-convex functions over strongly convex constraint sets, FW yields an improved convergence rate $O\left( \frac{1}{t} \right)$, which importantly does not hold for the $\ell_\infty$ constraint. 

\end{appendices}

\end{document}